# Full-reference image quality assessment-based B-mode ultrasound image similarity measurement

Kele Xu[1,2,3*], Xi Liu[2], Hengxing Cai[4,5], Zhifeng Gao[6]

[1]*Pierre and Marie Curie University, 75005 Paris, France*
[2]*ESPCI Paris, PSL Research University, CNRS, 1 rue Jussieu, Paris, 75005, France*
[3]*College of Electronic Science and Engineering, National University of Defense Technology, Changsha 410073, China*
[4]*School of Software and Microelectronics, Peking University, 100871 Beijing, China*
[5]*School of Engineering, Sun Yat-Sen University, 510006 Guangzhou, China*
[6]*School of Software and Microelectronics, Peking University, 100871 Beijing, China*

During the last decades, the number of new full-reference image quality assessment algorithms has been increasing drastically. Yet, despite of the remarkable progress that has been made, the medical ultrasound image similarity measurement remains largely unsolved due to a high level of speckle noise contamination. Potential applications of the ultrasound image similarity measurement seem evident in several aspects. To name a few, ultrasound imaging quality assessment, abnormal function region detection, etc. In this paper, a comparative study was made on full-reference image quality assessment methods for ultrasound image visual structural similarity measure. Moreover, based on the image similarity index, a generic ultrasound motion tracking re-initialization framework is given in this work. The experiments are conducted on synthetic data and real-ultrasound liver data and the results demonstrate that, with proposed similarity-based tracking re-initialization, the mean error of landmarks tracking can be decreased from 2 mm to about 1.5 mm in the ultrasound liver sequence.

**Keywords:** Full-reference image quality assessment, ultrasound image, motion tracking

## 1. Introduction

Ultrasound imaging technologies and ultrasound image interpretation have witnessed drastic progress since last decades. Compared to other image modalities (such as MRI, X-ray), the ultrasound imaging system is portable, non-invasive and relatively inexpensive. With the advances of the ultrasound machines, motion estimation (tracking) research in this area has increased remarkably[1,2,3]. Being a crucial component of motion estimation, tracking, noise filtering and ultrasound image registration, the local speckle pattern similarity measurement is of great interest for the researchers in ultrasound image interpretation. However, ultrasound-based tracking issue still poses great challenges. In brief, there are several factors influencing the performance of different proposed tracking algorithms[4]:

- Lack of stable features: for object tracking, extracting distinguished and robust feature is important to follow an object of interest. However, the widely-used features (such as intensity, texture) used in the optical image may not be suitable for ultrasound image. On the other hand, the robustness of the feature's quality is influenced by various factors, such as ultrasound probe orientation, acoustic shadows, and signal dropout.
- Large, swift and irregular deformation (motion): Optical flow is widely used to trace the object in the image sequence, which is based on the feature-points correspondence between adjacent frames. However, optical flow or its variants may not be applicable to the ultrasound images due to the speckle decorrelation. Moreover, deformation of the organs (such as liver and tongue) is large and the motion is quite fast, which poses an even

---

* Corresponding author: kelele.xu@gmail.com

greater challenge to track the motion using ultrasound imaging.

As mentioned earlier, despite sustainable efforts were made to improve the performance of tracking, tracking failure occurs frequently when different tracking approaches are applied to ultrasound image sequence, and manual tracking refinement is often needed. Nevertheless, since the trajectory of most of the organs' motion is repetitive (such as the tongue), if we can characterize the repetitiveness of the motion, tracking can be re-initialized automatically during the tracking process, which may be applicable to improving the performance of the tracking. To characterize the repetitiveness, a robust image similarity index is needed for the ultrasound image, which can be employed to measure the similarity between the reference image and other frames. Thus, the repetitiveness (or the periodicity) can be characterized. However, to date, not much attention has been paid to the ultrasound image similarity measurements, which can be used as a priori information to characterize the global motion. In fact, the potential applications of ultrasound image similarity measurement seem evident, which include ultrasound imaging quality assessment, abnormal function detection and motion tracking re-initialization.

The measurement of ultrasound image similarity can be considered as a full-reference image quality assessment (IQA) issue, which has witnessed notable progress during the last decades. In this work, we conduct a comparative study on different full-reference IQA methods for the ultrasound image similarity measurement task. Both synthetic and real ultrasound images are used for the comparison. Moreover, the application of image similarity-based motion tracking re-initialization is explored by using ultrasound liver data, and the results demonstrate that the tracking accuracy and robustness can be improved by using proposed automatic re-initialization method. The organization of the paper is given as follows: in section 2, we first make a brief technical summary on different full-reference IQA approaches. In section 3, a comparative study is conducted on the similarity measurement methods using both the synthetic image data and the real ultrasound data. Section 4 presents the proposed image-similarity based automatic re-initialization method in the ultrasound liver sequences, while the conclusion and future work are discussed in section 5.

## 2. Related work

Image quality evaluation is critical in many fields of image processing[5], which can be divided into objective and subjective evaluation. For objective measurement, according to the reference information used, it can be divided into three categories: full-reference IQA, reduced-reference IQA and no-reference IQA. The vast majority of IQAs are full-reference algorithms, which evaluate the quality of the distorted image based on the reference image. To calculate the ultrasound image similarity measurement, full-reference-based IQA is more desirable as the reference frame is available. Thus, in this paper, only full-reference IQA approaches will be discussed.

The simplest full reference IQA is the mean squared error (MSE), which calculates the mean of squared intensity differences between the input and reference image pixels.

Let $x_i, i=1,2,\cdots,N$ and $y_i, i=1,2,\cdots,N$ be the intensity values of gray images **X** and **Y** respectively. We can calculate MSE as:

$$MSE = \frac{1}{N}\sum_{i=1}^{N}(x_i - y_i)^2 \qquad (1)$$

Based on MSE, the peak signal-to-noise ratio (PSNR) is another index used to measure the similarity,

$$PSNR = 10\log_{10}(peakval^2 / MSE) \qquad (2)$$

where peakval is either specified by the user or taken from the range of the image datatype (the default value is set as 255).

Both MSE and PSNR aim to evaluate the local pixel-wise differences, and then convert these local measurements into a scalar which represents the overall quality[6]. Moreover, these measures assume that each pixel contributes equally during the evaluation.

As a recent thrust in full-reference image quality assessment, the structural similarity[7] (SSIM) outperforms the previous methods. The SSIM index measures three kinds of visual impact of changes in luminance, contrast and structure in an image. At each coordinate, the SSIM index is calculated within a local window. The overall index of similarity is a multiplicative combination of the three terms.

$$SSIM(x,y) = [l(x,y)]^\alpha [c(x,y)]^\beta [s(x,y)]^\gamma \qquad (3)$$

Based on the SSIM, a multi-scale version for the original SSIM was proposed[8], in which the correlation, luminance and contract measures are also applied to filtered and down-sampled versions of the image. At each scale $m$, the contract and structure terms are denoted as $c_m(x,y)$ and $s_m(x,y)$. The luminance term is computed only at the largest scale M, and denoted as $l_M(x,y)$. As a result, the overall quality evaluation is obtained by combining the measurements over scales.

$$MS-SSIM(X,Y) = [l_M(x,y)]^{\alpha_M} \prod_{m=1}^{M}[c_m(x,y)]^{\beta_i}[s_m(x,y)]^{\gamma_i} \qquad (4)$$

where, typically $M = 5$ and the exponents are elected such that and $\sum_{i=1}^{M} \gamma_i = 1$.

In brief, both the SSIM and MS-SSIM evaluate visual quality with a modified local measure of spatial correlation consisting of three components: mean, variance and cross-correlation. Various algorithms have been proposed based on the SSIM or MS-SSIM. Complex wavelet structural similarity (CW-SSIM)[9] is an extent of the SSIM method to the complex wavelet domain, which is a novel image similarity measurement due to its robustness to small distortions. To implement the CW-SSIM index for the comparison, the images are decomposed using a complex version of multi-scale, multi-orientation steerable pyramid decomposition[10, 11].

In more detail, to compute the CW-SSIM similarity between two ultrasound images, suppose we can represent the complex wavelet coefficients of the two frames by using $\mathbf{W}_X = \{w_{X,m} | m = 1,\ldots,M\}$ and $\mathbf{W}_Y = \{w_{Y,m} | m = 1,\ldots,M\}$, which are extracted at the same spatial location in the same wavelet sub-bands of the two images being compared, $M$ being the level of decomposition. Then we calculate the complex transform of them. The CW-SSIM similarity index between $\mathbf{X}$ and $\mathbf{Y}$ is:

$$CW-SSIM(\mathbf{X},\mathbf{Y}) = \frac{2\left|\sum_{m=1}^{M} w_{X,m} w_{Y,m}^*\right| + K}{\sum_{m=1}^{M} |w_{X,m}|^2 + \sum_{m=1}^{M} |w_{Y,m}|^2 + K} \quad (5)$$

where $w^*$ is the complex conjugate of wavelet coefficients $w$ and K is a small positive stabilizing constant.

Except for the structural-based similarity IQA, statistics-based IQA methods have also been explored. Visual information fidelity (VIF) is one important kind of those. In more detail, VIF algorithm aims to model the Human Visual System. Accordingly, the quality of the distorted image can be measured by using the statistical information which is shared between the reference image and the distorted image[13]. VIF is expressed as the ratio of sums of reference and distorted image information over different channels (output of a spatiotemporal kernel):

$$VIF = \frac{\sum_{j \in channels} I(\vec{C}^{N,j}; \vec{F}^{N,j} | s^{N,j})}{\sum_{j \in channels} I(\vec{C}^{N,j}; \vec{E}^{N,j} | s^{N,j})} \quad (6)$$

where

$$I(\vec{C}^N; \vec{E}^N | s^N) = \frac{1}{2} \sum_{i=1}^{N} \sum_{k=1}^{M} \log_2(1 + \frac{s_i^2 \lambda_k}{\sigma_n^2}) \quad (7)$$

$$I(\vec{C}^N; \vec{F}^N | s^N) = \frac{1}{2} \sum_{i=1}^{N} \sum_{k=1}^{M} \log_2(1 + \frac{g_i^2 s_i^2 \lambda_k}{\sigma_n^2 + \sigma_v^2}) \quad (8)$$

In the above expressions, $i$ is an index to the spatiotemporal block and $\lambda_k$ are the eigenvalues of $C_U$, the covariance matrix of a Gaussian random field vector U. The factors $s_i$, $g_i$, $\sigma_n^2$, $\sigma_v^2$ are parameters of the Gaussian Scale Mixture Model.

In addition to the aforementioned methods, there are numerous structural similarity measurements having been proposed for full reference IQA algorithms, and it's impossible to present the technical details of all algorithms exhaustively. On the other hand, due to the high frame-rate of ultrasound imaging and the natural fast movements (or deformations) during the tracking process, high calculation efficiency is also desirable while selecting the method. Thus, in this paper, only MSE, PSNR, SSIM, MS-SSIM, CW-SSIM and VIF are used to conduct the experiment on the synthetic data and the real ultrasound data.

## 3. Comparison between different full reference IQA methods

To make a quantitative comparison study between different similarity indices on the ultrasound image similarity measurements, we conducted an experimental study on the synthetic data from the perspectives on the speckle sensitivity and periodical local motion. Moreover, the ultrasound liver sequences are also used to in this section to evaluate the performance of different similarity indices.

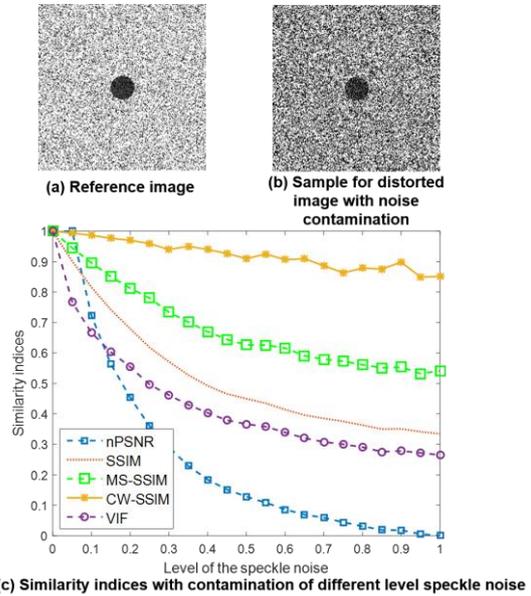

Fig. 1. Similarity indices with different levels of speckle noise contamination. Due to the fact that the ranges of some similarity indices are not suitable to be inserted into the figure, we normalized all of the indices into 0-1.

Inspired by the database construction method for visual quality evaluation[6], we use different distortions to contaminate the ground truth image. The comparison results obtained using synthetic data are given in Fig. 1, in which Fig. 1. (a) is the original synthetic image which will be regarded as the reference image (ground truth), Fig. 1. (b) is a sample frame contaminated by the multiplicative Rayleigh speckle noise of different level. Fig. 1. (c) gives the comparison between different indices over aforementioned speckle noise contaminations.

As can be seen from Figure 1, the PSNR, VIF, and SSIM indices cannot provide stable similarity descriptions with the occurrence of the speckle noise. MS-SSIM provides better performance with comparison to SSIM, while CW-SSIM has superior performance.

Another experiment is conducted to simulate the periodic motion in the ultrasound image sequence. In more detail, the periodic displacement is applied to the landmarks in the image. Different similarity indices were calculated on the synthetic sequence, which are shown in Fig. 2. As can be seen from Fig. 2, the CW-SSIM is able to characterize the periodic motion of the landmarks in the image sequence. To make a quantitative comparison between different indices, the Pearson correlation between the position of the landmarks and the similarity index is calculated for different similarity measurements, and the results are given in Table 1. As can be seen from the Table, the CW-SSIM provides higher correlation between the position of the landmark and the similarity index, which demonstrates that CW-SSIM can characterize the periodic motion better in the simulated image sequences.

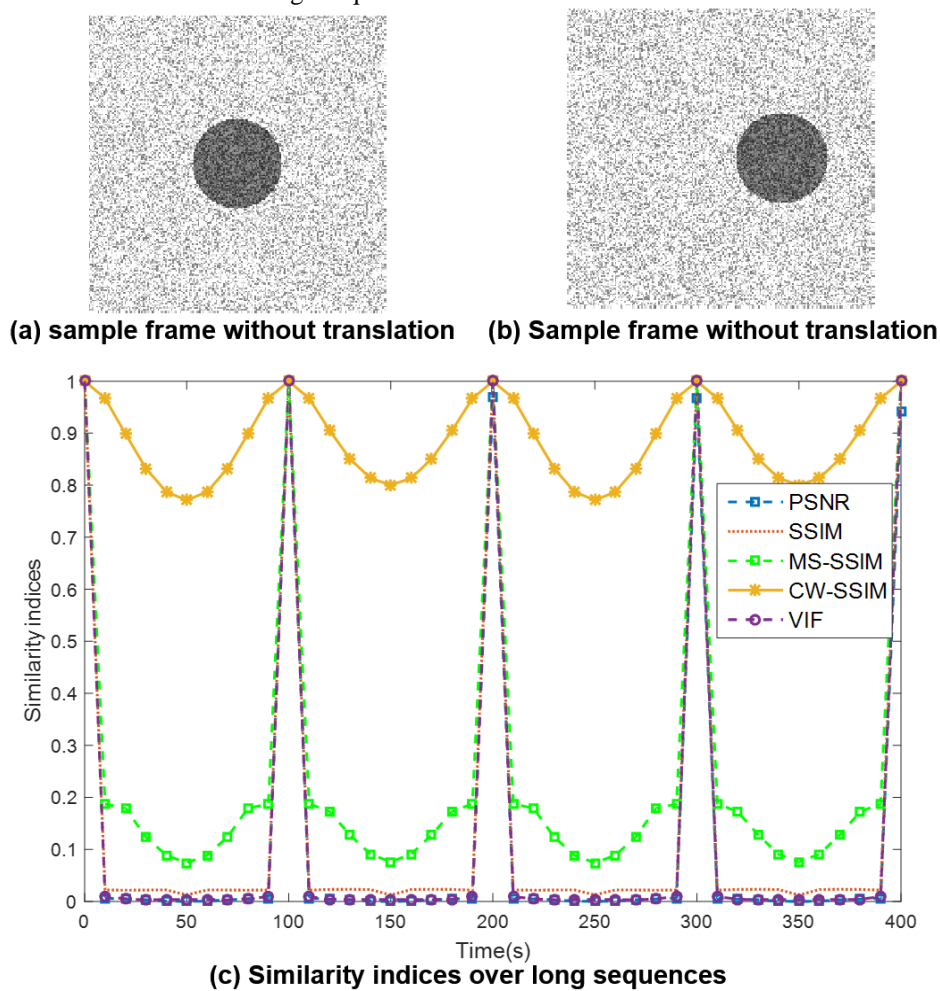

(a) sample frame without translation   (b) Sample frame without translation

(c) Similarity indices over long sequences

Fig. 2. Similarity indices of the simulated landmark in periodic motion. All indices are normalized into 0-1.

Table 1. Pearson Correlation between the position of the landmark and the image similarity index.

| Similarity index | Pearson Correlation (absolute value) |
|---|---|
| nPSNR | 0.6649 |
| SSIM | 0.6994 |
| MS-SSIM | 0.8017 |
| CW-SSIM | 0.9743 |
| VIF | 0.7078 |

To make a further comparison between different similarity indices on the real ultrasound data, the ultrasound liver sequences are employed to conduct the experiment. As shown in Fig. 3, the sample frame in the ultrasound liver data is given, in which the anatomical landmarks are labeled manually by domain experts (yellow point). The data were obtained using different ultrasound scanners and settings, and the images from two subjects had a range of spatial and temporal resolutions. Moreover, the length of the sequences varies from 1 second to 3 minutes, and the quality of the images was also highly variable.

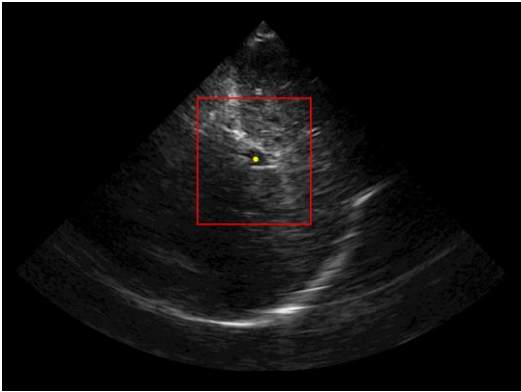

Fig. 3. Sample frame of in the dataset (for one subject). The yellow circle denotes the anatomical landmarks to track. The regions in the red rectangle are the regions of interest (ROI) selected manually for the similarity measurements.

The indices of different similarity measurements over the long sequence is given in Fig. 4, and the first frame is selected as the reference frame. To demonstrate the periodic synchronization between the similarity index and the displacement of landmarks, the lateral displacements of the hand-labeled landmarks are given in the two rows in Fig. 4, while the similarity indices are given in the bottom row. Similar to the experiment conducted on the synthetic data, to make a quantitative comparison between different indices by using the real data, the Pearson correlation between the position of the landmarks and different similarity index is calculated, which are given in Table 2. As can be seen from Table, the CW-SSIM provides superior performance.

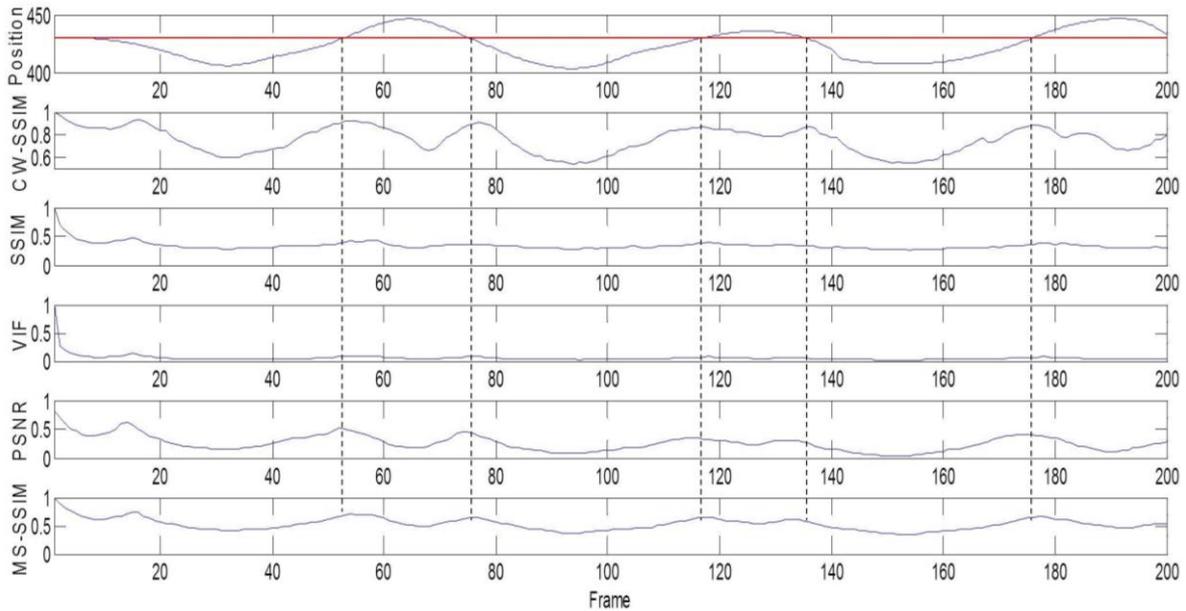

Fig. 4. The similarity indices over an ultrasound liver sequence.

Table 2. Pearson Correlation between the position of the landmark and different image similarity index in the ultrasound liver sequences.

| Similarity index | Pearson Correlation |
| --- | --- |
| nPSNR | 0.1143 |
| SSIM | 0.0609 |
| MS-SSIM | 0.1511 |
| CW-SSIM | 0.2002 |
| VIF | 0.0557 |

In Fig. 4, it is worthwhile to note that: when the landmark goes back to the position which was labeled manually in the first frame (or reference frame), the CW-SSIM index will reach its local maximum, which demonstrated the superior performance of CW-SSIM over the other similarity indices. As the result, we will employ the CW-SSIM as the ultrasound image similarity measurement to explore the potential application of full reference IQA in the ultrasound motion tracking issue, which is given in next section.

**4. Full reference IQA-based ultrasound motion tracking re-initialization**

In this section, we will explore the potential applications of the ultrasound image similarity measurement. In more detail, we explore the potential applications in motion tracking re-initialization. As aforementioned in the introduction part, robust and accurate motion tracking and tissue motion analysis in the ultrasound sequences can be greatly helpful for the image-guided intervention and therapy. Specifically, in this article, we aim to address the tracking failure issue in the ultrasound liver sequences using the global similarity measurement.

Despite of sustainable efforts being made, motion tracking in the ultrasound liver image sequences has been posing a significant challenge due to the noise, shadows, signal dropouts, out-of-plane motion and speckle decorrelation. To address the aforementioned tracking failure problem, in this work, we explore the use of image similarity-based tracking reset method, which makes use of the periodicity of respiratory motion in the ultrasound liver sequences. The goal of the proposed method is to explore re-initializing the tracking automatically, thus improving the robustness and accuracy of the tracking method. Although we aim to address the issue of tracking failure recovery in the ultrasound liver sequences, it may also be applied to other organs' tracking problem using ultrasound imaging[14].

Unlike the general tracking problem in the video sequence, this automatic reset method is feasible in the ultrasound liver sequences as the breathing is periodic. The proposed reset method works as follows. Before anatomic landmarks tracking is carried out, the similarity coefficient is calculated, between current frame and the reference frame. If this coefficient exceeds a set threshold, the positions of anatomical landmarks are reset to the initial positions. This provides a method to prevent accumulation of errors over long sequences, which can lead to erroneous tracking, and amounts to a sort of "automatic reset" of the tracking starting points based on initial a priori information.

To demonstrate the feasibility of the proposed re-initialization method, we incorporate the re-initialization method into two widely used tracking methods in the liver sequences. One is based on the 2D normalized cross-correlation (NCC) algorithm; the other approach uses the Mean Shift-based tracking. In the NCC-based method, a region of interest (ROI) is defined around each annotation in the first ultrasound frame. In subsequent ultrasound frames a larger search region is defined. The position of maximum correlation from the NCC algorithm is used to identify the new position of the ROI. In the mean shift-based method, the rescaling and orientation procedures are applied.

The tracking algorithms (NCC and Mean Shift-based methods) were developed in MATLAB Release 2015a. Experiments were conducted to track the landmarks in the 2 volunteer B-mode ultrasound liver sequences. The results were evaluated by comparison with manual annotations of liver feature (vessels) throughout each image sequence which was provided after automated tracking was complete. Tracking accuracy was evaluated using the Euclidean distance between tracked points and manually annotated points, which was summarized by the mean and standard deviation. The results are shown in Table 3. The mean error for all ROIs in sequences group are about 1-1.5 mm with the proposed automatic reset method, while the error is 2 mm without the automatic reset method.

Table 3. Tracking error comparison with and without automatic re-initialization.

| Sequence | Mean error and standard deviation of different tracking methods | | | |
| --- | --- | --- | --- | --- |
| | NCC | | Mean-Shift | |
| | Without reset | With reset | Without reset | With reset |
| 1 | 1.75 ± 1.23 | 1.69 ± 0.96 | 2.48 ± 1.51 | 2.04 ± 0.96 |
| 2 | 3.71 ± 3.84 | 3.01 ± 1.71 | 2.42 ± 1.83 | 1.75 ± 1.33 |
| ALL | 2.73 ± 2.54 | 2.35 ± 1.34 | 2.45 ± 1.67 | 1.90 ± 1.15 |

Moreover, we extend our automatic tracking re-initialization method to reset the tongue contour tracking issue in the ultrasound sequence, on which we obtain superior performance with the proposed method with comparison to the results obtained without reset[15,16].

## 5. Conclusion and future work

In this paper, we made a comparison of the structural indexes in the ultrasound image sequences. Moreover, we also give an automatic tracking reset method in ultrasound liver sequences of long duration. The proposed method utilizes the periodicity of motion of the liver. The experimental results demonstrate that proposed method can improve the accuracy and robustness of the tracking. On the other hand, it can be easily extended to 3D ultrasound tracking failure issue.

For future work, we will explore incorporating the multiplicative speckle noise model into the similarity indices, improving the processing speed and extending the proposed method to other organs' tracking problem in the ultrasound image sequences.